\documentclass{article}

\PassOptionsToPackage{numbers, compress}{natbib}

\usepackage[final]{neurips_2023}

\usepackage[toc,page]{appendix}
\usepackage{subfig}
\usepackage{graphicx}
\usepackage{float}
\usepackage{amsmath}

\newcommand{\modelname}{TCNCA}
\newcommand{\camready}[1]{\textcolor{black}{#1}}

\usepackage[utf8]{inputenc} 
\usepackage[T1]{fontenc}    
\usepackage{hyperref}       
\usepackage{url}            
\usepackage{booktabs}       
\usepackage{amsfonts}       
\usepackage{nicefrac}       
\usepackage{microtype}      
\usepackage{xcolor}         
\usepackage{multirow}       

\title{TCNCA: Temporal Convolution Network with Chunked Attention for Scalable Sequence Processing}

\author{Aleksandar Terzić$^{1,2}$\thanks{Research conducted at IBM Research -- Zurich.}\,\,, 
Michael Hersche$^{1,2}$,
Geethan Karunaratne$^{1}$\\
\textbf{Luca Benini}$^{2}$,  
\textbf{Abu Sebastian}$^{1}$,
\textbf{Abbas Rahimi}$^{1}$\thanks{Corresponding author: abr@zurich.ibm.com} \\
{
$^{1}$IBM Research -- Zurich, $^{2}$ETH Zurich
}
}

\begin{document}

\maketitle

\begin{abstract}
MEGA is a recent transformer-based architecture, which utilizes a linear recurrent operator whose parallel computation, based on the FFT, scales as $O(LlogL)$, with $L$ being the sequence length.
We build upon their approach by replacing the linear recurrence with a special temporal convolutional network which permits larger receptive field size with shallower networks, and reduces the computational complexity to $O(L)$. 
The resulting model is called \textbf{\modelname{}}, a \textbf{T}emporal \textbf{C}onvolutional \textbf{N}etwork with \textbf{C}hunked \textbf{A}ttention. 
We evaluate \modelname{} on EnWik8 language modeling, long-range-arena (LRA) sequence classification, as well as a synthetic reasoning benchmark associative recall. 
On EnWik8, \modelname{} outperforms MEGA, reaching a lower loss with $1.37\times$/$1.24\times$ faster forward/backward pass \camready{during training}. 
The dilated convolutions used in \modelname{} are consistently and significantly faster operations than the FFT-based parallelized recurrence in GPUs, making them a scalable candidate for handling very large sequence lengths: they are up to $7.07\times$/$2.86\times$ faster in the forward/backward pass for sequences up to 131\,k.
Further on LRA, \modelname{} achieves, on average, $1.28\times$ speed-up during inference with similar accuracy to what MEGA achieves. 
%
On associative recall, we find that even a simplified version of \modelname{}, without excessive multiplicative and additive interactions, remains superior or competitive to MEGA on a \camready{range of sequence lengths and vocabulary sizes}.
%

\end{abstract}

 \section{Introduction}
\label{sec:intro}
The Transformer \cite{transformer} is a powerful class of neural networks which has found success in a variety of tasks including image processing \cite{vit}, physical system modeling \cite{tfphysics}, drug discovery \cite{drugformer}, but perhaps most notably, language modeling \cite{bert}, \cite{t5}, \cite{gpt}. 
While undeniably a strong candidate for a universally applicable neural network, the operator at its backbone, \emph{attention}, faces some crucial limitations. 
We consider two limitations, including the $O(L^2)$ computational and memory complexity~\cite{eff_tf_survey} of attention, as well as its poor performance in long sequence classification, namely on the long-range-arena (LRA) dataset~\cite{lra}, where it is drastically outperformed by \emph{linear recurrent models}~\cite{s4,resurrecting,s5}; however, these models lag behind the transformer on language modeling~\cite{h3}. 
A more extensive review of related works can be found in Appendix A.
%
%

A recent neural network, MEGA \cite{mega}, combines the strengths of \emph{linear recurrences} and \emph{attention} in a manner which scales sub-quadratically. Concretely, MEGA combines the damped exponential moving average (EMA) known from time-series analysis \cite{dampema}, with chunked attention which operates on fixed-size non-overlapping blocks in the input sequence. It achieves scores competitive with the state-of-the-art in a range of disparate tasks including language modeling on the EnWik8 dataset~\cite{enwik8} and LRA sequence classification~\cite{lra}.
%

We focus on EMA, which maps $\mathbf{x_t}\in \mathbb{R}^h$ to $\mathbf{y_t} \in \mathbb{R}^h$ using the parameters $\mathbf{\alpha, \delta} \in [0,1]^h, h \in \mathbb{N}_+$ as:
\begin{equation}
\label{eq:dampema}
    \mathbf{y_t = \alpha \odot x_t + (1 - \alpha \odot \delta) \odot y_{t-1}}.
\end{equation}
This operation can be directly computed as per equation \ref{eq:dampema}. 
However, during training and non-causal data processing, it can equivalently be computed as a convolution with a kernel which is of the same shape as the input data \cite{mega}. This convolution can be efficiently performed in $O(LlogL)$ time in the frequency domain \cite{parallelizingLMU}, \cite{s4}. This mode of operation is interesting because it allows for a higher utilization of GPUs' parallel processing capabilities ~\cite{parallelizingLMU}. 



In this work, we investigate the performance and runtime effects of replacing the bottleneck EMA within the MEGA processing stack with a dedicated temporal convolutional neural network (TCN)~\cite{tcn_vs_lstm, bytenet,wavenet,eegtcnet}, an operator which scales linearly with the sequence length.
The TCN employs dilated convolutions, which allow the network to achieve a large receptive field with few parameters. 
TCNs are typically implemented as a cascade of \emph{residual blocks}, in which each block applies two dilated convolution operations with equal dilations. 
In order to quickly reach large receptive fields, the dilation exponentially increases with each successive block~\cite{tcn_vs_lstm,eegtcnet}. Our model differs from what is usually used in literature in that it only includes a single dilated convolution operation per residual block.
This construction allows for a larger receptive field size with shallower networks. Details are given in Appendix E.
We call the resulting model, which combines a TCN with chunked attention, \modelname. 

We find that on EnWik8 language modeling, \modelname{} outperforms MEGA~\cite{mega} (and Transformer-XL~\cite{tfxl}), achieving a BPC score of 1.01, in addition to $1.37\times$/$1.24\times$ faster forward/backward pass.  
\camready{On a synthetic reasoning benchmark, \emph{associative recall}, a simplified version of \modelname{} (see Appendix C) is competitive with MEGA over a range of different sequence lengths and vocabulary sizes.}
%
On 64-dimensional sequences of lengths ranging from 8192 to 131072, the employed dilated convolution operator is up to $7.07\times$ and $2.86\times$ faster than the parallelized EMA of MEGA in the forward and backward pass, respectively.
This signifies the scalability of the approach to long sequences thanks to its linear complexity.
On the LRA classification tasks, \modelname{} slightly underperforms MEGA by only 0.1\% on average, while achieving $1.28\times$ inference speedup. 
%
%
%

\section{The TCNCA model}
\label{sec:method}

\begin{figure}[b]
  \centering
  \includegraphics[width=0.9\textwidth]{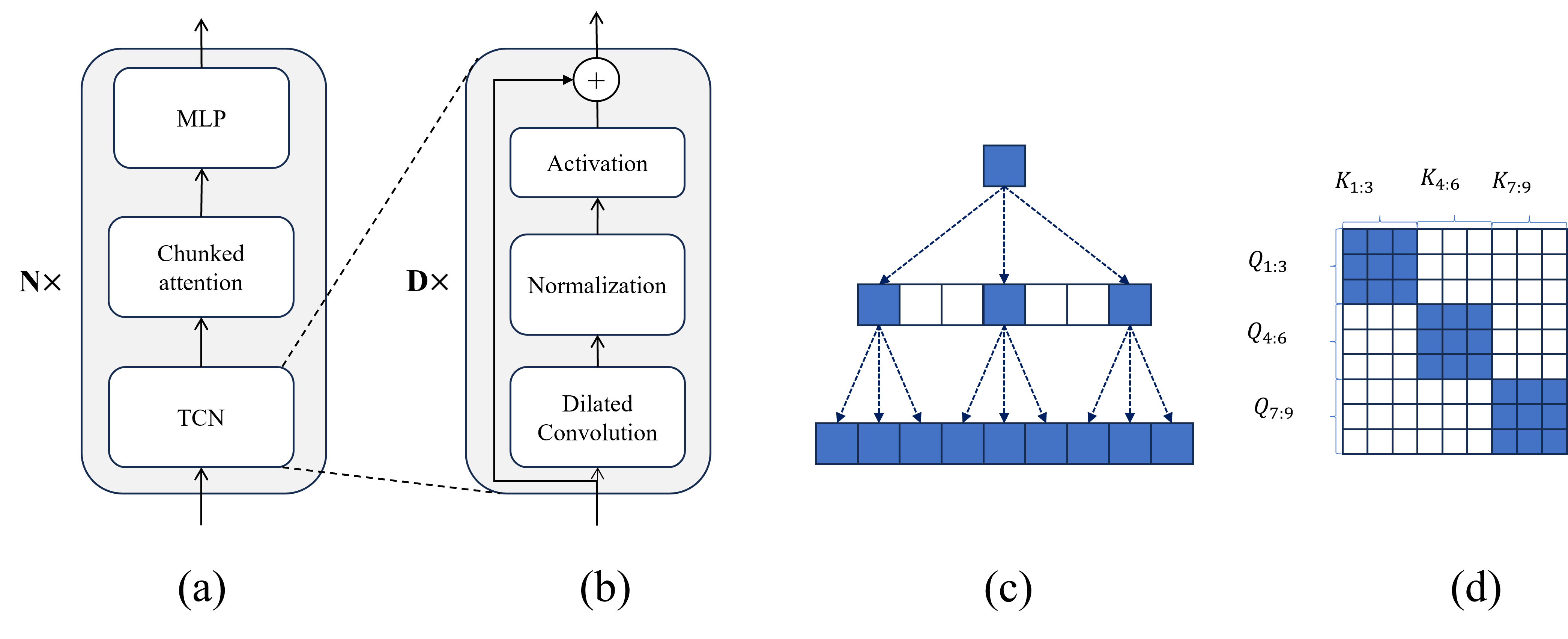}
  \caption{(a) Simplified high-level overview of the \modelname{} model. (b) The TCN residual block. (c) Connectivity of a TCN with kernel size $K=3$, dilation factor $f=3$, and depth $D=2$. (d) Chunked attention operation which computes query-key similarities in fixed-width non-overlapping windows, shown with chunk size 3.}
  \label{fig:main_sketch}
\end{figure}

An overview of the model and the operations used therein is shown in Figure \ref{fig:main_sketch}. At a high-level, the model can be thought of as a concatenation of a temporal convolutional neural network (Figure~\ref{fig:main_sketch}b) with chunked attention (Figure~\ref{fig:main_sketch}d). 
The sketch is simplified; the actual construction follows the one defined by MEGA \cite{mega}, and is outlined in Appendix C.

Figure~\ref{fig:main_sketch}a shows a depth-$N$ sequence processing stack. Each of the $N$-many layers consists of a temporal convolutional network and chunked attention, both of which operate along the time axis, followed by a multi-layer perceptron (MLP) operating along the feature axis. For each embedding dimension, a TCN with its own set of trainable parameters is instantiated.

The TCN block in Figure~\ref{fig:main_sketch}a is expanded in Figure~\ref{fig:main_sketch}b. Three integer hyperparameters govern the TCN construction; kernel size $K$, dilation factor $f$, and depth $D$. The TCN consists of $D$-many residual blocks, each of which implements a dilated convolution operation whose dilation is determined by the layer index $i=0,...,D-1$ and $f$ as $f^i$. 
In Figure~\ref{fig:main_sketch}c, we show the connectivity pattern of a TCN with $D=2$, $f=3$ and $K=3$.



Following the TCN, which scales as $O(L)$, we have chunked attention. As already noted, it computes the query-key similarities only within fixed-size non-overlapping windows within the sequence, as shown in Figure~\ref{fig:main_sketch}d. This is also an $O(L)$ operation.

\section{Experiments}
\label{sec:experiments}

\paragraph{EnWik8 language modeling}
EnWik8 is a dataset which comprises a subset of the English Wikipedia. 
We train and evaluate our model on EnWik8 character-level language modeling in the same manner as was done in MEGA~\cite{mega}. The results are shown in Table~\ref{tab:enwik8}. More details are given in Appendix F.

\begin{table}[!h]
\caption{EnWik8 bit-per-character scores. Results marked with a star (*) are taken from~\cite{mega}.}
\label{tab:enwik8}
\centering
\begin{tabular}{l|lll}
\hline
Model & Transformer-XL & MEGA & \modelname{} \\ \hline
BPC   & 1.06*           & 1.02* & \textbf{1.01}      \\ \hline
Parameters & 41M & 39M & 39M \\ \hline
\end{tabular}

\end{table} 

\modelname{} outperforms the Transformer-XL \cite{tfxl} as well as MEGA \cite{mega}, reaching a 1.01 BPC score. For transparency's sake, we have to note that the scores reported in relevant literature are rounded down to 2 digits after the decimal point, hence we do the same. With 4 digits after the decimal point, the score we achieve is 1.0144 BPC.

We measure the forward and backward pass speed-up on a 16GB Nvidia V100 GPU during training. 
During training, \modelname{} achieves a $\mathbf{1.373\times}$ speed-up in the forward pass and a $\mathbf{1.245\times}$ speed-up in the backward pass, compared to MEGA.
However, speeding up the inference runtime of the generative tasks is not straightforward and is one of the limitations of this work (see Appendix B).

\paragraph{Long-range-arena}
Long-range-arena~\cite{lra} comprises six classification tasks with sequence lengths ranging from 1024 to 16384. The benchmarks are varied, including pattern detection, sentiment classification, mathematical reasoning, and visual logical reasoning. 
We use the same dimensionalities, hyperparameters, and attention chunk sizes as those used in MEGA~\cite{mega}, and select the TCN construction as per Appendix D. Results are shown in Table~\ref{table:lra}.

\begin{table}[b]
\caption{Long-range-arena accuracies (\%) of state-of-the-art models. The Transformer scores are taken from the reproduction in MEGA~\cite{mega}. All other results, excluding \modelname{}, were taken from the respective papers. The last row reports the end-to-end inference speed-up of \modelname{} measured against MEGA-chunk.}
  \label{table:lra}
  \centering
\begin{tabular}{llllllll}
\hline
Model             & ListOps & Text    & Retrieval & Image   & Path    & Path-X  & Average \\ \hline
Transformer \cite{transformer} \cite{mega} & 37.1 & 65.2 & 79.1 & 42.9 & 71.8 & 50 & 57.7 \\ \hline
S4D \cite{s4d} & 60.5 & 86.2 & 89.5 & 89.9 & 93.1 & 91.9 & 85.2 \\ \hline
S5 \cite{s5}& 62.2 & 89.3 & 91.4 & 90.1 & 95.3 & 98.6 & 87.8 \\ \hline
LRU \cite{resurrecting}& 60.2 & 89.4 & 89.9 & 89.0 & 95.7 & 96.0 & 86.7 \\ \hline
SGConv \cite{whatmakes} & 61.4 & 89.2 & 91.1 & 87.97 & 95.4 & 97.8 & 87.1 \\ \hline
MEGA chunk \cite{mega} & 58.7 & 90.2 & 91.0 & 85.8 & 94.4 & 93.8 & 85.6 \\ \hline
\modelname  & 59.6 & 89.8 & 89.4   & 86.8 & 94.5 & 92.7 & 85.5 \\ \hline
Speedup (forward pass) & $1.05\times$  & $1.25\times$   & $1.18\times$  & $1.24\times$ & $1.25\times$ & $1.73\times$ & $1.28\times$ \\ \hline
\end{tabular}
\end{table}

Although \modelname{} lags behind the state-of-the-art state space method, S5~\cite{s5}, by 2.3\%, it is on par with MEGA-chunk (just an average of a 0.1\% lower accuracy) while achieving an average inference speed-up 28\%. 

\paragraph{Associative recall}
This synthetic benchmark requires faithful attention and measures the basic reasoning capability of neural sequence models, remembering associations between pairs of tokens \cite{arecall} \cite{h3}. For example, given a sequence of tokens \emph{a 2 c 4 b 3 d 1}, if the model is prompted with \emph{a}, the expected output is \emph{2}, the token following \emph{a} in the input sequence. If it were prompted with \emph{b}, the correct output would be \emph{3}, etc.

%
\camready{As mentioned, \modelname{} is based on MEGA~\cite{mega}, and as such it involves an intricate interconnection between the different modules it is composed of. We report \modelname{} scores for the associative recall in a setting in which the module interconnection is significantly simplified by eliminating excessive multiplicative and additive interactions (\modelname{}-simple, see Appx. C).}
%
%
\camready{Over the investigated range of vocabulary sizes and sequence lengths in Table~\ref{tab:as_rec_voc_10}, \modelname{}-simple remains competitive with MEGA.}
%

\begin{table}[]
\caption{Associative recall accuracy (\%) with varying sequence lengths and vocabulary sizes.}
\label{tab:as_rec_voc_10}
\centering
\begin{tabular}{l|cc|cc}
\hline
& \multicolumn{2}{c|}{Vocabulary size 10} & \multicolumn{2}{c}{Vocabulary size 20} \\ \hline
Seq. len. & MEGA & TCNCA-simple & MEGA & TCNCA-simple \\ \hline
64    & 98.8 & 100  & 62.4 & 56 \\ \hline
1024  & 99.6 & 100  & 99.4 & 97.6 \\ \hline
4096  & 100  & 100  & 100 & 99.6 \\ \hline
8192  & 98.2 & 100  & 98.6 & 99.2 \\ \hline
\end{tabular}
\end{table}

\begin{figure}[H]
  \centering
  \subfloat[Forward pass runtime measurements.]{\includegraphics[width=0.48\textwidth]{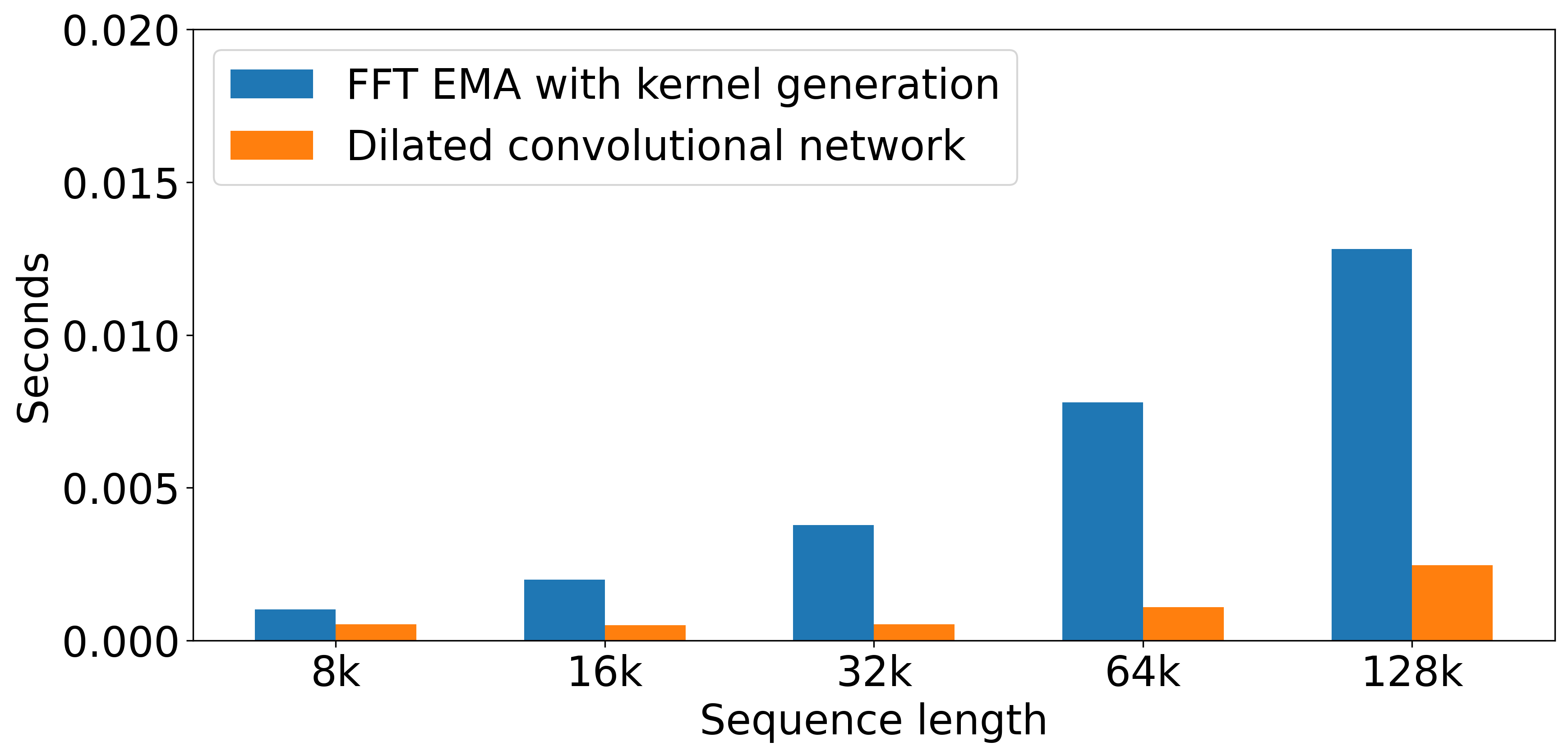}\label{fig:f1}}
  \hfill
  \subfloat[Backward pass runtime measurements.]{\includegraphics[width=0.48\textwidth]{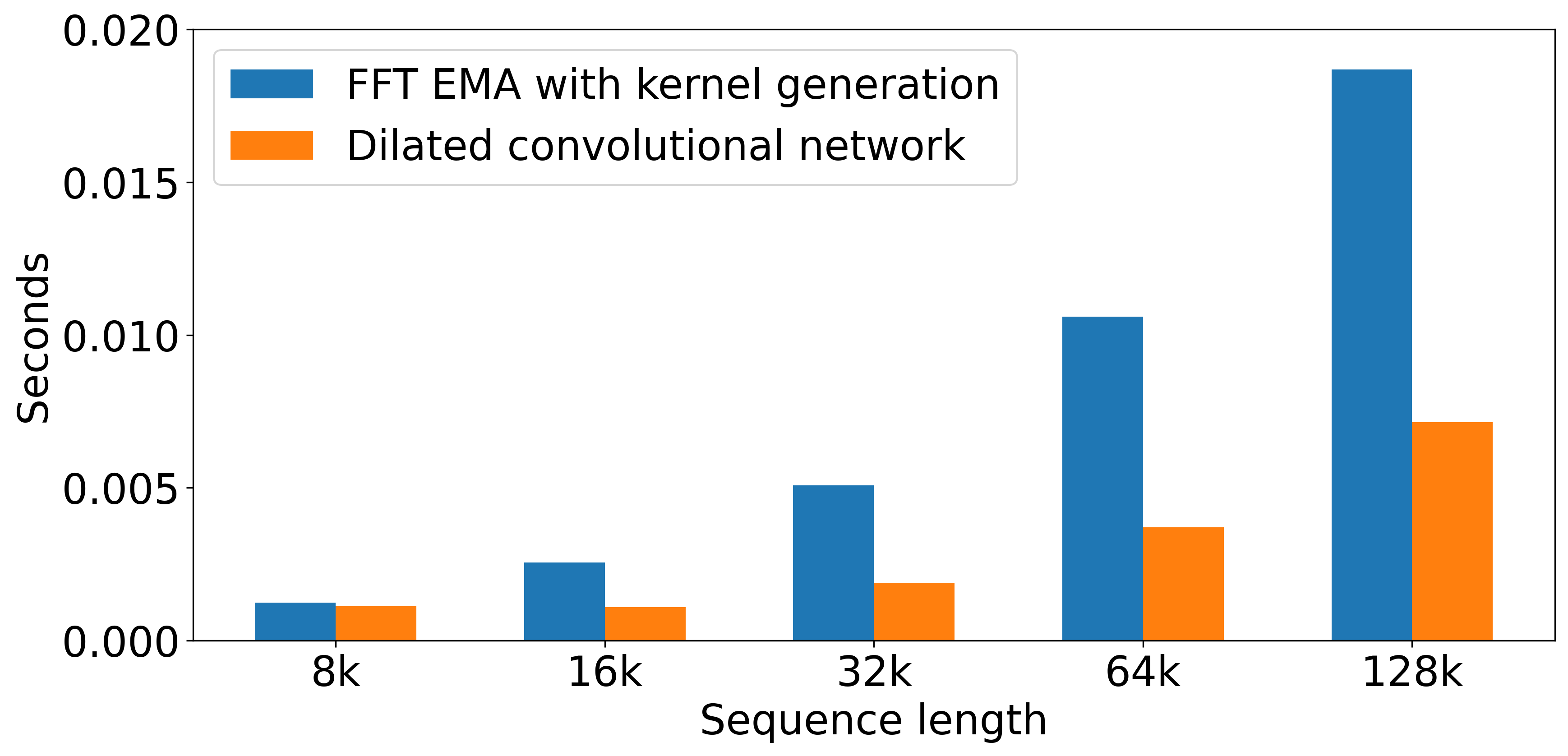}\label{fig:f2}}
  \caption{Run-time comparisons between a parallel linear recurrence including kernel generation (blue) and a dilated CNN (orange) for the forward and backward pass, with varying sequence lengths. The dilated convolutional network is consistently the faster operation.}
  \label{fig:runtimes}
\end{figure}

\paragraph{Parallelized EMA vs. dilated convolution runtime measurements}
We measure the forward and backward-pass runtimes of a dilated convolutional network and a parallelized EMA recurrence over a range of sequence lengths\camready{, and report the results in Figure \ref{fig:runtimes}}. For a clear comparison of the two operations, we strip both of them of residual connections, non-linearities as well as normalization layers.
They are roughly parameter-matched, with EMA having 64 parameters and the dilated convolution having 68 parameters.
The dilated convolutional network is configured with $K=17$, $D=4$, and $f$ is increased until the receptive field of the network is larger than the sequence length it operates on. The benchmarks were run on an Nvidia V100 with 16\,GB of VRAM. Further details are given in Appendix H.

\section{Conclusion}
\label{sec:conclusion}

In this work inspired by ground-breaking results from the team behind MEGA~\cite{mega}, we show that a TCN and chunked attention hybrid model, \modelname{}, is able to compete with the state-of-the-art models on Enwik8 language modeling and Long-Range-Arena sequence classification.
During training and non-causal inference workloads, \modelname{} consistently exhibits inference speed-ups in the range of $5\%$ to $73\%$ compared to MEGA-chunk.
%
\camready{We show that a simplified version of \modelname{} solves the \emph{associative recall} synthetic reasoning benchmark with a similar accuracy as does MEGA.}
Finally, we show that on the Nvidia V100 GPU, a dilated convolutional network is consistently faster than an FFT-based parallelized EMA recurrence over a wide range of sequence lengths. 
Some of the limitations of our approach are detailed in Appendix B.

\bibliography{bibliography}

\newpage

\appendix

\section{Related work}
\label{apx:related}

Long sequence modeling is a rich area of research within the deep learning community. A non-exhaustive list of work relevant to ours is outlined below.

\textbf{Linear recurrent models}

A particularly prominent family of linear recurrent models comes in the form of linear state-space models (LSSMs).
In their original formulation, linear state-space models are based on the HiPPO framework \cite{hippo}, in which an optimal mechanism for incrementally updating a fixed-size state while processing online streams of information by projecting onto an orthogonal polynomial basis is derived.
Based on the HiPPO framework, the seminal work S4 \cite{s4} introduces a new addition into the neural sequence processing family of operators, linear state-space models. S4 is, to the best of our knowledge, the first method to significantly advance the state-of-the-art on LRA classification \cite{lra}. 
Many other linear state-space models follow; S4D \cite{s4d} diagonalizes the linear recurrence, GSS \cite{gss} introduces a gating mechanism for improving LSSMs' performance on language modeling, S5 \cite{s5} introduces MIMO LSSMs.

We had mentioned that linear state-space models, in their typical formulation, are not suitable for language modeling. H3 \cite{h3}, motivated by contemporary work on mechanistic interpretability \cite{ICL}, proposes a multiplicative interconnection of linear state-space models which they find significantly improves their performance on this task.

One of the newer additions to the family of linear recurrent models is the \emph{LRU} model \cite{resurrecting}, which steps away from the state-space model framework which is based on discretizing an implicit continuous state-space model while still achieving near-state-of-the-art accuracies on LRA.


\textbf{Long convolutional models}

We denote models which apply convolutions whose kernels are of the same length as the input sequence as \emph{long convolutional models}. 
SGConv \cite{whatmakes} constructs a sparsely parametrized long kernel with an exponentially decaying structure and finds that this achieves strong performance on LRA. Hyena hierarchy \cite{hyena} achieves impressive scores on language modeling with a long convolutional kernel, without attention. It is to the best of our knowledge the highest performing attention-free language model. 
FlashButterfly \cite{flashbutterfly} explores simple long convolution kernel constructions that are effective on LRA classification and furthermore develops a hardware-aware algorithm for efficient computation of FFT-based long convolutions.

\textbf{Linear complexity attention}

A rich area of research addresses the quadratic scaling of the attention operation by deriving more scalable approximations thereof. A survey of such approaches can be found in \cite{eff_tf_survey}.

\textbf{TCNs for sequence modeling}

CDIL-CNN employs circular dilated convolutions on several datasets including Long-Range-Arena \cite{cdilcnn}. WaveNet \cite{wavenet} is a generative audio model based on the TCN architecture. ByteNet \cite{bytenet} employs the TCN for machine translation.

\textbf{TCN-Attention hybrid models}

TCAN~\cite{tcan} employs a cascade of dilated convolutions with full self-attention. Their architecture scales quadratically with the sequence length, has one convolution per decoder stack, and the causal attention masking they implement, described in Section 2.3, is in fact not causal. TConvTransformer~\cite{tconvtf} describes a similar idea, a quadratic complexity concatenation of a TCN and multi-head self-attention, and is evaluated on 3D human pose and shape estimation.

\section{Limitations}
\label{apx:limitations}

On generative tasks, an important example being language modeling, linear recurrent models offer the unique advantage of parallel training and recurrent inference. A temporal convolutional neural network is a naturally parallel operation, but it is not trivial to adapt it to efficiently operate on generative tasks. In a generative mode of operation, we would need to cache intermediate computations generated by the TCN and re-use them at later time steps. We have not implemented this mode of operation in our work, opting instead to demonstrate the feasibility and applicability of the idea. Implementing this mode of operation is left for future work.


Another advantage of recurrent models lies in the fact that their receptive field size is theoretically unlimited. The TCN's receptive field size is restricted through the construction that we impose by defining the kernel size, network depth, and dilation.


The TCN introduces three new hyperparameters, namely kernel size, dilation factor, and network depth. This significantly expands the hyperparameter search space.

There exists at least one alternative option for parallelizing linear recurrent systems, the \emph{parallel scans} algorithm \cite{parallelScans}. We have not compared the runtimes of the TCN with this algorithm.

As hinted at in the method description section in the main text, the chunked attention module is in fact significantly more intricately interconnected with the TCN than Figure 1 makes it seem like. The full construction is shown in Appendix \ref{apx:attn_details}. Future work should investigate the necessity of each part of the module with the goal of simplifying the construction and maximizing its performance.

\section{Detailed overview of the attention module}
\label{apx:attn_details}

\begin{figure}[]
  \centering
  \includegraphics[width=1.0\textwidth]{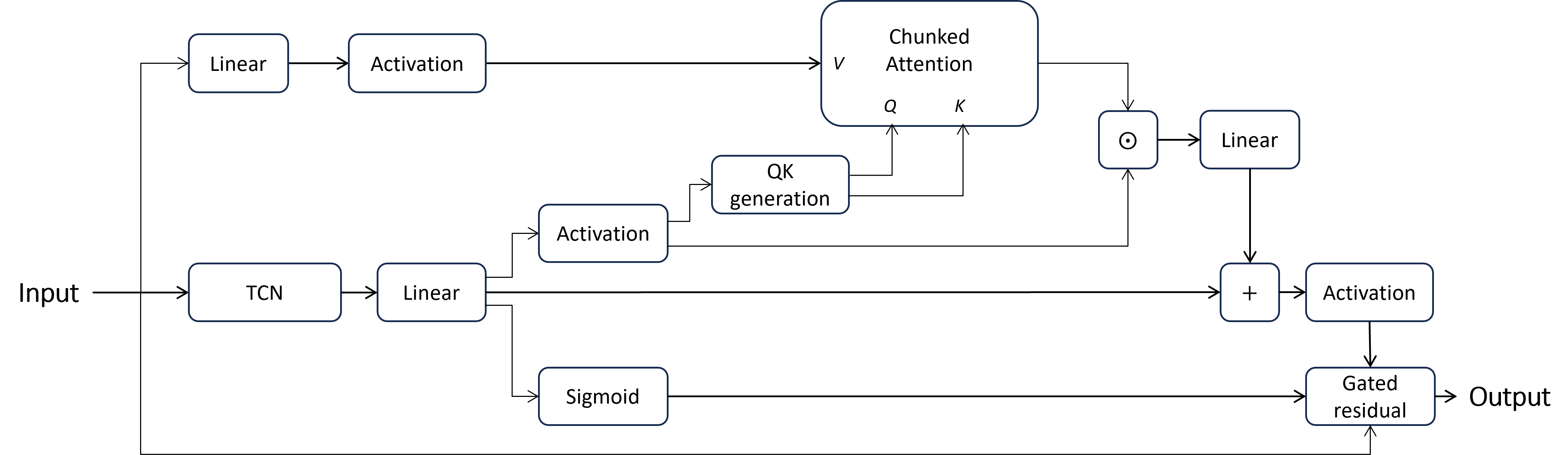}
  \caption{The module interconnection used in \modelname{} is inherited from MEGA and is significantly more complex than the simplified sketch we have shown in the main text. It involves several multiplicative and additive interactions between outputs of different operators. QK generation applies a diagonal linear transformation on the input data and adds a trainable bias to it, with separate parameters for queries and keys. The gated residual is equivalent to the construction defined in Highway Networks \cite{highway}.}
  \label{fig:gate_attn_overview}
\end{figure}

The attention module is more intricately interconnected with EMA/TCN than Figure 1 in the main text would make it seem. Details are given in Figure \ref{fig:gate_attn_overview}.

\begin{figure}[]
  \centering
  \includegraphics[width=1.0\textwidth]{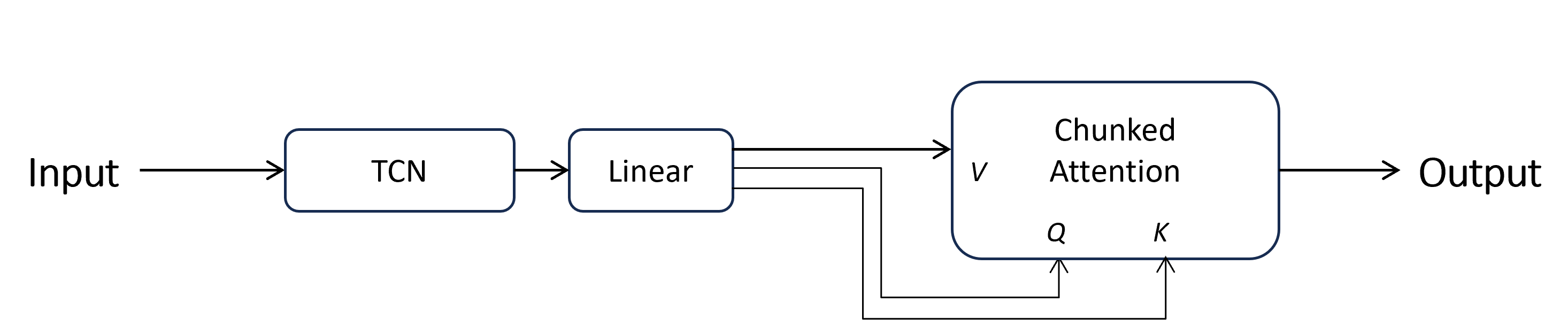}
  \caption{\modelname{}-simple is a significantly simpler version of the full \modelname{} model, and is used in the associative recall benchmark.}
  \label{fig:simple_tcn:attn}
\end{figure}

For the \emph{associative recall} benchmark, we used a simpler construction, called \modelname{}-simple, which is outlined in Figure\ref{fig:simple_tcn:attn}

\section{Hyperparameter selection}
\label{apx:hyperparams}

For LRA and EnWik8, the TCN structure, which is defined by the kernel size, dilation factor, and TCN depth, was selected by measuring inference run-times of a range of configurations, selecting a range of those which exhibit a speed-up, and finally training networks with those structures on the given dataset. It is a rather costly procedure, since there exist many different TCN structures which exhibit a speed-up, and they differ between benchmarks. A more principled way of selecting TCN configurations should be investigated in future work.

For experiments on the long-range-arena dataset, the same learning rate scheduler, learning rate, weight decay, dropout values, activation functions, and vector dimensionalities as those used in MEGA \cite{mega} were used.

For experiments on the EnWik8 dataset, the same learning rate scheduler, dropout values, activation functions and vector dimensionalities as those used in MEGA \cite{mega} were used. A learning rate of 0.005 and a weight decay of 0.2 were used to obtain the best score.

The only varied hyperparameters in the associative recall experiments are dropout and the learning rate. Dropout is selected from the small grid defined by two points, [0, 0.1]. Learning rate was swept over [1e-5, 1e-4, 1e-3, 1e-2, 1e-1].

\section{TCN construction}
\label{apx:tcn_constr}

\begin{figure}[h!]
\centering
  \includegraphics[width=0.8\textwidth]{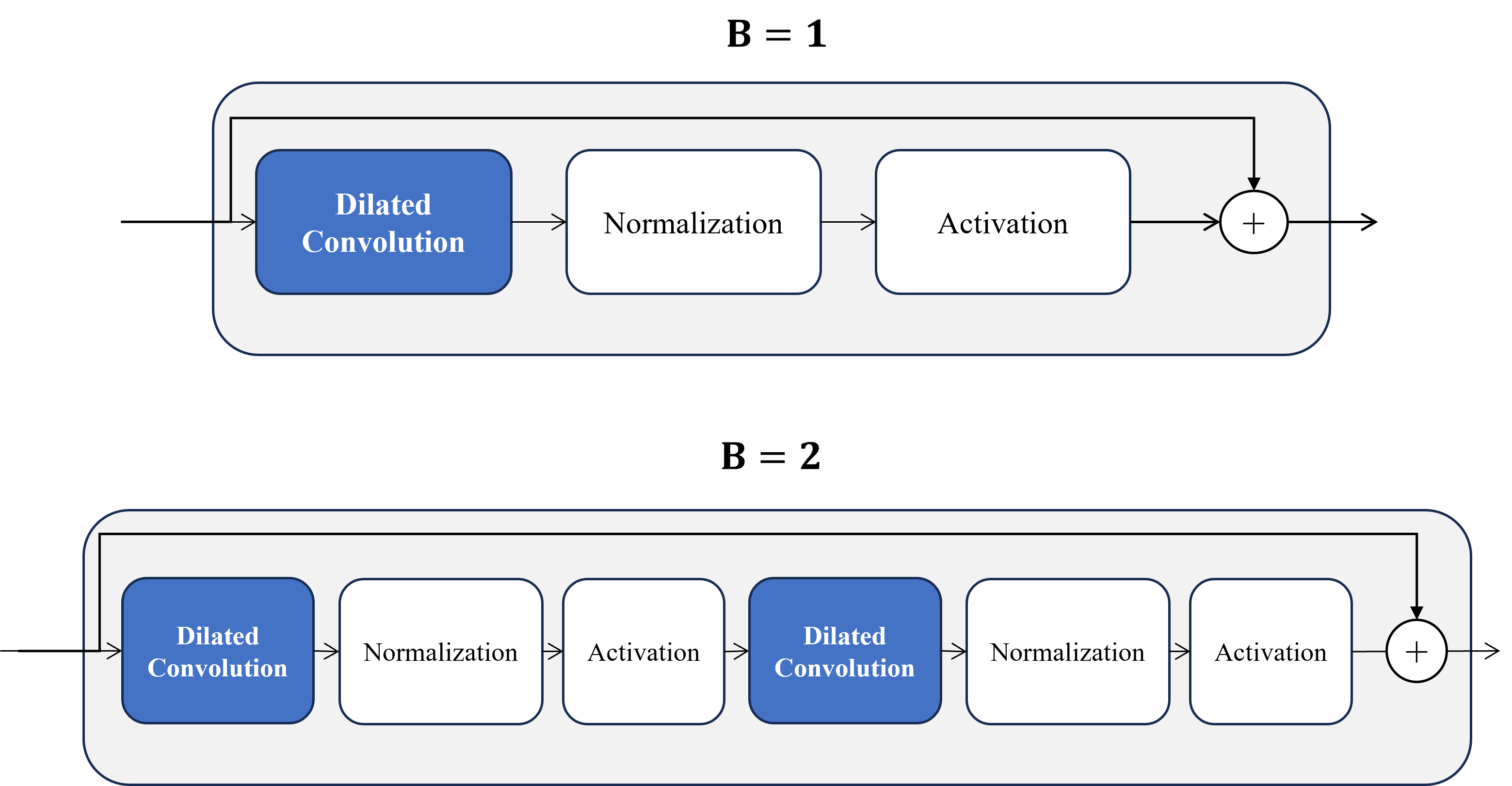}
  \caption{The hyperparameter $B$ controls the number of dilated convolutions with equal dilation within a single residual block. The full network still consists of $D$-many such blocks. On the top, we show a residual block with $B=1$, which is what we used in our experiments. $B=2$ is another popular option in published works on TCNs.}
  \label{fig:tcn_blockdepth}
\end{figure}

In addition to the TCN hyperparameters introduced in the main text, these being kernel size $K$, dilation factor $f$, and depth $D$, we will in this section introduce a new hyperparameter, $\mathbf{B}$, which stands for the number of dilated convolutions within \emph{a single residual block}. This is explained in Figure \ref{fig:tcn_blockdepth}.

The receptive field size of the TCN can be calculated as $1 + B * (K-1) * \frac{f^D-1}{f-1}$. It scales exponentially with $D$ and linearly with $B$. Keeping the total number of dilated convolution operations $D\times B$ equal, the largest receptive field size is achieved when $B$ is minimized and $D$ is maximized, which is the reason we opted for $B=1$ in our experiments.

\section{EnWik8 train and test details}
\label{apx:enwik8}

The data is split into consecutive chunks of size 2048, which is also what the attention chunk size is set to. At training time, we randomly load 2, 3, or 4 consecutive chunks of text to be processed by the model. During evaluation, the attention chunk size is set to 4096, and 3 consecutive chunks of text are loaded. We train for 250 epochs. Just as in MEGA \cite{mega}, attention is augmented with relative positional bias, the particular one used in this task being RoPE \cite{roformer}.

\section{Associative recall setup}
\label{apx:arecall}

We must note that, while $N$ from Figure 1 (a) is 2, $D$ from Figure 1 (b) must be larger than 1 in order to obtain a large enough receptive field of the TCN. Hence, our model does consist of a depth-2 sequence decoding stack as was used in \cite{h3}, but each of the decoder layers uses a deeper TCN, typically of depth 3 or 4.

For sequence length 64, we use embedding dimension 32 and an attention chunk size of 32, corresponding to the standard quadratic self-attention. For all other sequence lengths, an embedding dimension of 128 with a chunk size of 128 is used.


\section*{The need for self-attention}
\label{apx:need_attn}

In this section, we demonstrate the performance difference between attention-less models and those which hybridize EMA/TCN with attention on EnWik8 (Tables \ref{table:mega_enwik8_role_of_attn} and \ref{table:tcn_enwik8}) and LRA (Table \ref{table:tcn_ffn_lra}). On LRA, we are in fact able to achieve strong performance using a TCN-based attention-free model, outperforming the best TCN-based attention-free model known in the literature, CDIL-CNN \cite{cdilcnn}.

\begin{table}[!h]
\centering
\caption{The role of attention in MEGA on EnWik8 after full training. The hyperparameters used in MEGA were re-used in the EMA-MLP experiment set. The result marked with a star (*) is taken from \cite{mega}.}
\label{table:mega_enwik8_role_of_attn}
\begin{tabular}{lc}
\hline
 & Enwik8 loss $\downarrow$ \\ \hline
EMA-MLP stack & 1.96 \\ \hline
MEGA & 1.02* \\ \hline
\end{tabular}

\end{table}

\begin{table}[!h]
\centering
\caption{Introducing chunked attention after the TCN significantly reduces the BPC loss on enwik8. Both models went through a hyperparameter grid search but were trained for a limited number of epochs, hence the gap between the result reported here and in the main text.}
\label{table:tcn_enwik8}
\begin{tabular}{lc}
\hline
 & Enwik loss after 250k training steps \\ \hline
TCN-MLP stack & 1.44 \\ \hline
\modelname & 1.08 \\ \hline
\end{tabular}

\end{table}

\begin{table}[!h]
\centering
\caption{Comparison of S5 \cite{s5}, CDIL-CNN \cite{cdilcnn}, and our implementation of a dilated convolutional neural network on several tasks from the long-range-arena dataset. Our implementation outperforms the best-known TCN-based result from literature, the CDIL-CNN work \cite{cdilcnn}, on all four LRA benchmarks which we both evaluate on.}
\label{table:tcn_ffn_lra}
\begin{tabular}{lcccccc}
\hline
 & ListOps & Text & Retrieval & Image & Path & Path-X  \\ \hline
 S5 \cite{s5}& 62.1\% & 89.3\% & 91.4\%  & 88.0\% & 95.3\% & 98.6\%  \\ \hline
CDIL-CNN \cite{cdilcnn} & --- & 87.6\% & 84.3\% & 64.5\% & 91.0\% & --- \\ \hline
Our TCN-MLP & 56.9\% & 89.6\% & 85.9\%  & 91.4\% & 97.2\% & 91.6\% \\ \hline
\end{tabular}

\end{table}

\section{Runtime benchmark methodology}
\label{apx:bm_setup}

The EMA hidden dimension (dimension expansion from \cite{mega}, Section 3.1) is set to 8. Within the multi-dimensional damped EMA algorithm presented in MEGA \cite{mega}, this results in 64 parameters. The TCN is always of depth $D=4$ with 17 parameters in each layer, hence it consists of 68 parameters, slightly more than the amount present in EMA. The dilation factor is increased until the TCN's receptive field size is greater than or equal to the sequence length.

All operations were implemented in the PyTorch 2.0.1 framework. The run-time measurements were obtained using the PyTorch benchmark module \footnote{https://pytorch.org/tutorials/recipes/recipes/benchmark.html}. LRA and EnWik8 run-times were measured on a 32\,GB Nvidia V100 GPU. All other run-times were measured on a 16\,GB Nvidia V100 GPU.

\section{Effect of EMA kernel generation on inference runtime}
\label{apx:kergen_cost}

\begin{figure}[!h]
  \centering
  \includegraphics[width=1.0\textwidth]{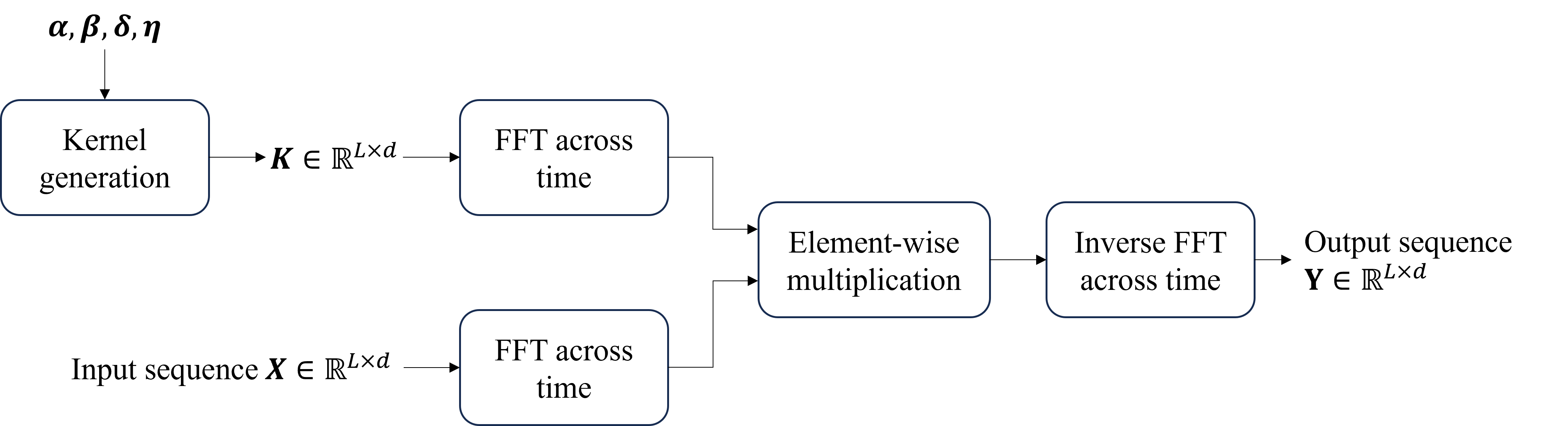}
  \caption{Computing the EMA linear recurrence in parallel invokes the computational pipeline shown in this figure. The kernel computation is given in MEGA \cite{mega}, Appendix A. There exists at least one alternative way of computing the long convolution, using the \emph{parallel scans} algorithms \cite{parallelScans}, which we did not consider in this work.}
  \label{fig:fft_ema_pipeline}
\end{figure}

\begin{figure}[H]
  \centering
  \includegraphics[width=0.6\textwidth]{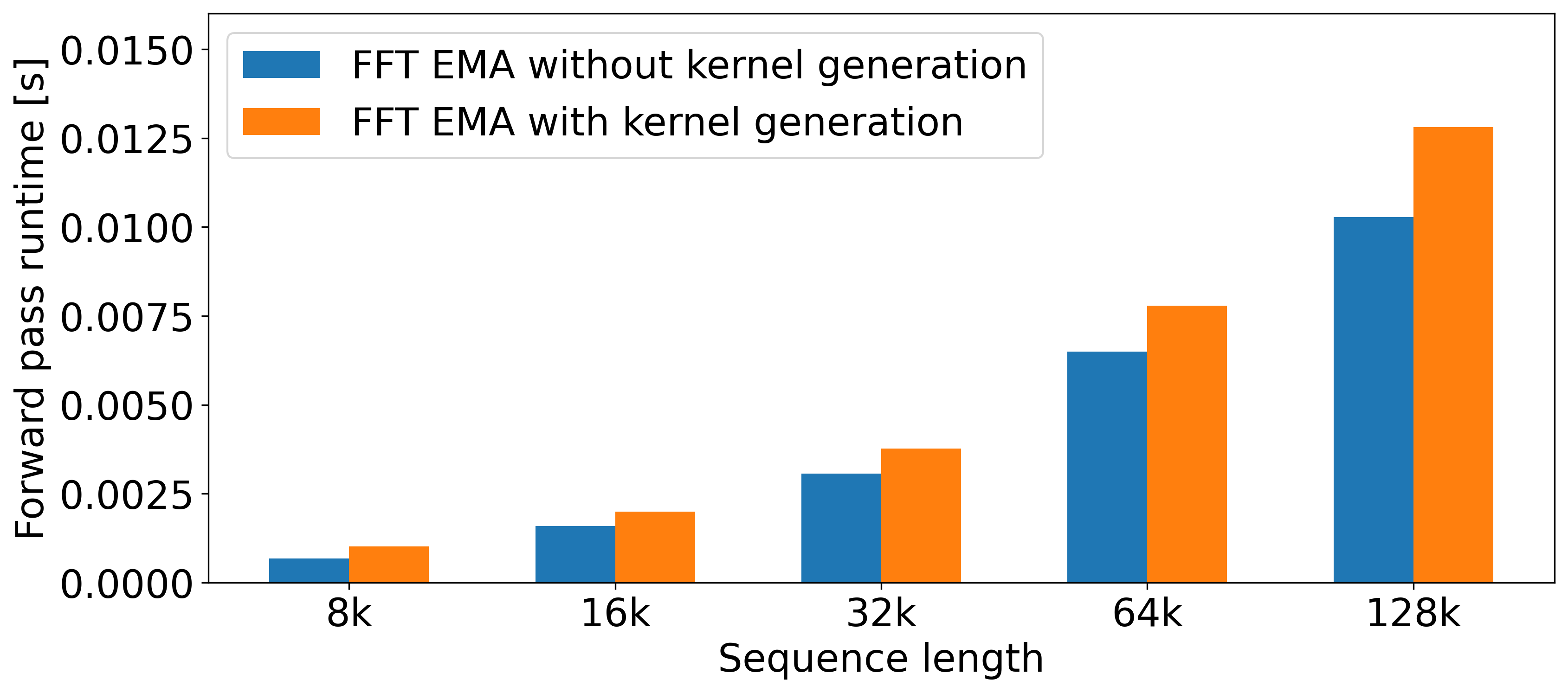}
  \caption{Comparing the forward-pass runtimes between the FFT-based parallel EMA with and without kernel generation (see Figure \ref{fig:fft_ema_pipeline}).}
  \label{fig:kergen_cost}
\end{figure}

The runtime measurements presented in the main text include the kernel generation cost. This is certainly necessary during training, but at inference, one might consider storing very long kernels and truncating them based on the input sequence length. 
This would reduce the inference runtimes of FFT-based EMA convolution. Runtime comparisons of FFT-based EMA convolutions with and without kernel generation are shown in Figure \ref{fig:kergen_cost}. Speed-ups of the version without kernel generation vs. the version that includes kernel generation are given in Table \ref{table:nokergen_speedup}.

\begin{table}[!h]
\centering
\caption{Speedup of FFT-EMA without vs. with kernel generation.}
\label{table:nokergen_speedup}
\begin{tabular}{llllll}
\hline
        & 8k    & 16k  & 32k   & 64k   & 128k  \\ \hline
Speedup & $1.51\times$ & $1.25\times$ & $1.23\times$ & $1.20\times$ & $1.25\times$ \\ \hline
\end{tabular}
\end{table}

\end{document}